\documentclass[conference]{IEEEtran}
\usepackage{cite}
\usepackage{amsmath,amssymb,amsfonts}
\usepackage{booktabs,multirow,tabularx,graphicx}
\usepackage{xcolor}
\usepackage[hidelinks]{hyperref}
\usepackage{tikz}
\usetikzlibrary{arrows.meta,positioning,calc}
\newcommand{\doop}{\operatorname{do}}

\begin{document}
\title{Instructions Before Actions: Prospective and Retrospective Language Supervision for Driving Vision--Language--Action Data}
\author{\IEEEauthorblockN{Angel Martinez-Sanchez}
\IEEEauthorblockA{\textit{University of California, Merced} \\}
\and
\IEEEauthorblockN{Ross Greer}
\IEEEauthorblockA{\textit{University of California, Merced} \\}
}

\maketitle

\begin{abstract}
Driving vision--language--action (VLA) models can accept navigation instructions, yet predicting a driver's trajectory does not establish that a model follows the instruction. We study how driving-language supervision is collected and evaluated, distinguishing \emph{prospective} instructions issued before an action from \emph{retrospective} labels written after observing it. We develop ADVLAT-Engine to synchronize spoken instructions, road-scene video, and vehicle trajectories, and compare three navigation apps on 48 replayed routes. On 87 maneuvers announced by both Google Maps and a passenger, the passenger uses more scene-dependent language. We evaluate four open driving VLAs before and after fine-tuning on prospective navigation-app drives. Fine-tuning reduces passenger-drive trajectory error by 23--59\%, but instruction effects remain inconsistent. At real intersections, counterfactual instructions reveal stronger responsiveness: fine-tuned Alpamayo~2 Super gains 31 and 18 accuracy points at 15 and 5~m, while Qwen-Drive follows alternative maneuvers more reliably with a discrete command than with a sentence. We additionally collected retrospective annotations from four people for the driving recordings and describe controlled comparisons of annotation content, timing, and downstream learning; those analyses are pending. Our findings separate maneuver prediction from instruction-following and establish a framework for comparing prospective and retrospective supervision.
\end{abstract}

\section{Introduction}
Driving vision--language--action (VLA) models combine visual understanding with trajectory planning, and several accept natural-language navigation instructions~\cite{shao2024lmdrive,renz2025simlingo,alpamayo15,alpamayo2}. In principle, this enables a passenger to direct a vehicle instead of simply accepting a precomputed route. Yet an accurate trajectory prediction does not show that the model used the instruction. Driving benchmarks can be dominated by recent ego motion~\cite{li2024ego,zhai2023rethinking}, and recent work reports instruction insensitivity in driving and robot VLAs~\cite{hu2026lcs,fei2025liberoplus,you2026language2drive}.

An important distinction lies in \emph{when} training language was created. Most driving-language datasets attach instructions or descriptions \emph{retrospectively}, after a trajectory has been observed~\cite{roy2024doscenes,arai2024covla,qian2024nuscenes,sima2024drivelm}. But in deployment, a passenger or navigation app provides the instruction \emph{prospectively}, before the maneuver. This type of data may describe the same eventual turn, but they differ in temporal availability, natural lead time, and information available to the speaker. A retrospective annotator can see how the maneuver ends, but a prospective navigator cannot. This matters when a model must decide \emph{when} an instruction applies, rather than merely identify which maneuver occurred.

Prospective data need not require expensive manual annotation. Navigation apps continuously issue spoken directions before upcoming maneuvers. Recording those prompts alongside camera observations and GNSS trajectories produces naturally timed instruction--vision--action examples. Passengers also provide a complementary source, often saying ``right here'' or referring to nearby objects instead of road names or metric distances. These distinctions matter for applications beyond conventional turn-by-turn navigation as after arriving at a destination, a passenger might ask an autonomous vehicle to move to a particular curb or building entrance. Such requests require grounding language in the observed scene and planning a safe, permitted maneuver.

\begin{figure*}[t]
\centering
\includegraphics[width=\textwidth]{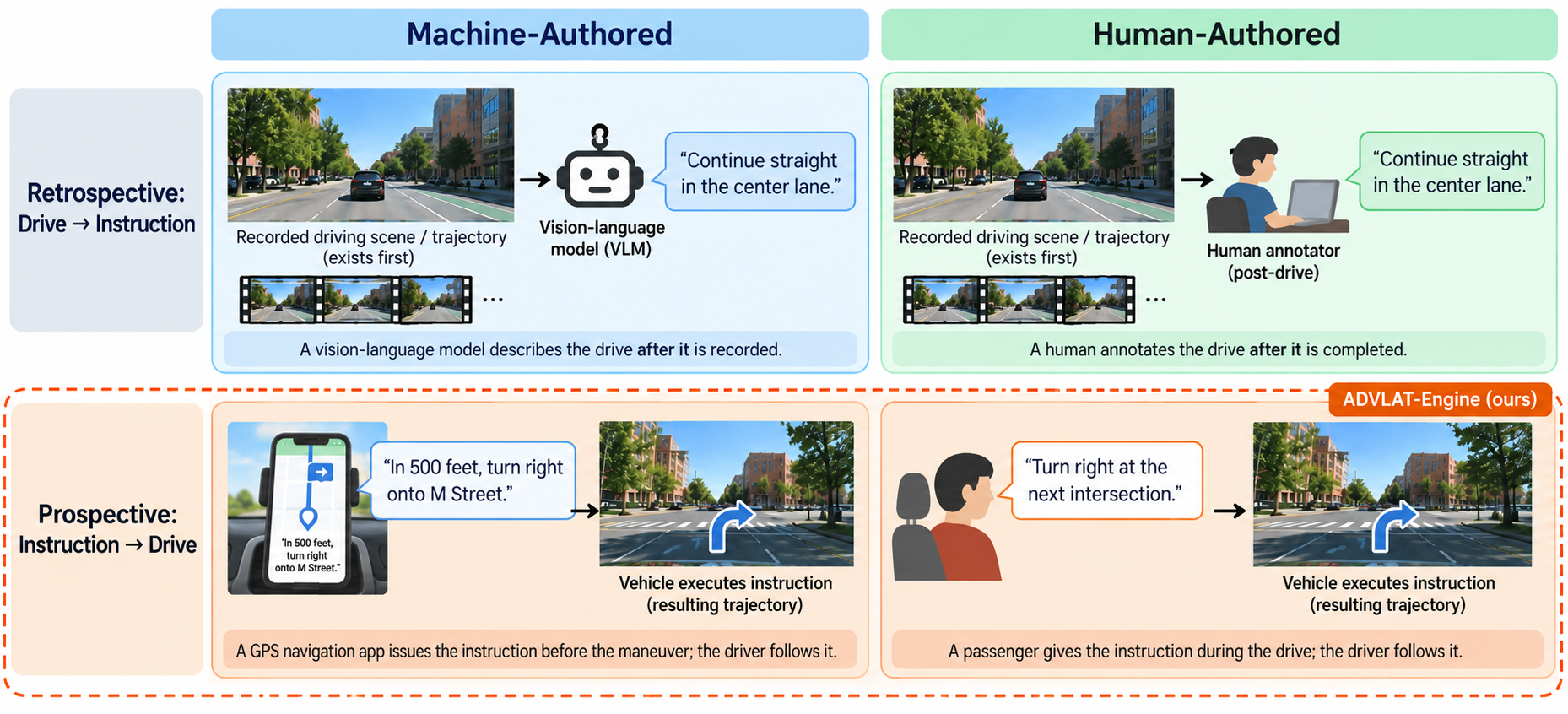}
\caption{Instruction sources differ in \emph{authorship} (machine or human) and \emph{timing} (retrospective or prospective). Retrospective labels are created after observing a completed drive; navigation apps and passengers issue prospective instructions before the relevant action. ADVLAT-Engine collects the latter alongside synchronized vision and trajectories. Temporal order motivates, but does not by itself identify, causal instruction effects.}
\label{fig:taxonomy}
\end{figure*}

We bring these perspectives together in one study. We compare navigation-app language, record prospective app and passenger instructions with Autonomous Driving Vision-Language-Action Triad Engine (ADVLAT-Engine), and test whether four driving VLAs trained on app recordings actually respond to passenger language. We also collected retrospective annotations from four people for the same driving videos, allowing us to compare instructions given before the maneuver with those written after observing it, while holding the recorded driving scene and trajectory fixed. Figure~\ref{fig:taxonomy} summarizes the supervision sources. The existing experiments establish trajectory and instruction-following results; we specify the retrospective-annotation comparisons as planned analyses rather than report results that have not yet been computed.

Our contributions are:
\begin{enumerate}
    \item A source-and-timing taxonomy for driving language supervision.
    \item ADVLAT-Engine and a controlled study of three navigation apps, alongside matched app and passenger instructions.
    \item A benchmark of four driving VLAs using paired instruction removal and counterfactual tests at real intersections.
    \item A retrospective annotation corpus from four annotators, with a proposed evaluation framework for comparing prospective and retrospective supervision.
\end{enumerate}

\section{Related Work}
\label{sec:related}
\subsection{Language-Driven Navigation and Conditional Driving}
Early language-based navigation systems connect verbal goals or landmarks to embodied action. LM-Nav~\cite{shah2023lm} uses language and visual observations to navigate toward landmarks, whereas conditional imitation learning~\cite{codevilla2018conditional} explicitly separates steering behavior by high-level maneuver commands. These examples show why the ability to choose among legal actions at the same intersection matters: predicting the most likely road-following trajectory is not sufficient for following a new route instruction. More recent language-conditioned driving systems include LMDrive~\cite{shao2024lmdrive}, SimLingo~\cite{renz2025simlingo}, Alpamayo~1.5 and 2~\cite{alpamayo15,alpamayo2}, Qwen-Drive~\cite{zhou2026qwendrive}, and Impromptu~VLA~\cite{chi2025impromptu}. Their architectures and instruction interfaces differ, motivating matched instruction-removal and counterfactual evaluation rather than comparisons based only on displacement error.

\subsection{Driving-Language Datasets and Temporal Supervision}
Talk2Car~\cite{deruyttere2019talk2car} pairs nuScenes scenes~\cite{caesar2020nuscenes} with free-form commands. nuScenes-QA~\cite{qian2024nuscenes}, DriveLM~\cite{sima2024drivelm}, doScenes~\cite{roy2024doscenes}, and CoVLA~\cite{arai2024covla} offer complementary forms of driving-related language, with differences in whether text is an instruction, description, or answer to a question. Learning to Drive (L2D)~\cite{huggingfaceblog} constructs navigation-style language from previously logged routes. The important distinction for our study is not simply whether language accompanies a trajectory, but whether it was \emph{available before the maneuver} or \emph{derived after observing it}. Post-hoc language can be valuable supervision, but an annotator may know the executed turn and has no naturally observed instruction lead time. Prospective app and passenger recordings retain when a command was given, including windows well before a turn enters a short prediction horizon; that lead time is relevant for variable-horizon planning~\cite{roy2026doplan}. We do not claim that all prior driving-language datasets have the same annotation mechanism; the timing and intended semantics must be considered individually.

\subsection{Retrospective Annotation, Hindsight Relabeling, and Efficiency}
Automatic annotation for autonomous driving has often focused on geometric labels: OpenAnnotate2~\cite{zhou2024openannotate2}, 3D-BAT~\cite{zimmer20193d}, and ActiveAnno3D~\cite{ghita2024activeanno3d} help reduce effort in labeling scene structure. Language creates a related but different problem. In robot learning, hindsight experience replay~\cite{andrychowicz2017hindsight} and instruction augmentation~\cite{xiao2023robotic} associate completed trajectories with goals or descriptions that they satisfy. These strategies expand training data by exploiting outcomes, and need not produce bad policies. Yet they can differ from a navigator's original request in specificity, temporal placement, and information available to the speaker. ADVLAT-Engine is complementary: it automates collecting already-spoken instructions rather than synthesizing hindsight labels. Our four-annotator corpus permits a same-video comparison of these mechanisms instead of assuming one source is universally better.

\subsection{Causal Confusion and Evaluating Instruction-Following}
Conditioned policies can exploit predictive shortcuts without learning the intended control variable. De Haan \emph{et al.}~\cite{dehaan2019causalconfusion} study causal confusion in imitation learning and the value of targeted interventions while Codevilla \emph{et al.}~\cite{codevilla2019limitations} analyze generalization issues in driving behavior cloning. Closed-loop imitation learning also suffers distribution shift because a policy changes the observations it later receives~\cite{ross2011dagger}. Recent driving-language evaluations compare removal or alteration of an instruction~\cite{martinezsanchez2026natural,renz2025simlingo,hamid2026icrdrive,you2026language2drive}. These motivate two distinct measurements in our work: whether removing passenger text changes trajectory error, and whether changing the requested legal maneuver changes the \emph{correct} predicted maneuver at an otherwise identical recorded intersection. Neither test alone establishes how the human driver's original action was caused, but both test instruction dependence in the model.

\section{Causal Problem Formulation}
\label{sec:formulation}
\subsection{Two Processes Hidden by the Same Training Tuple}
Let $S_t$ be the driving context available immediately before an instruction becomes active, including camera observations, recent ego-motion history, road layout, and any observed route context. Let $L$ be an instruction spoken prospectively; $A$ is the subsequent driving trajectory. Let $L'$ denote a retrospective label authored after the maneuver is recorded and $U$ collect unobserved common causes, such as the driver's intended destination or preferences not fully represented in $S_t$. The same training representation $(S,\text{language},A)$ may therefore arise from different data-generating processes~\cite{pearl1995causal,pearl2000causality}.

For prospective supervision, $L$ can precede and influence $A$, while context and latent intent can influence both. For retrospective supervision, $A$ is recorded before $L'$ is written, and the completed action can inform the label. A crucial distinction is the \emph{speaker's} data-generating process versus the \emph{learned policy}: that a past label could not cause a past drive does not imply a model trained on that label cannot later respond to it.

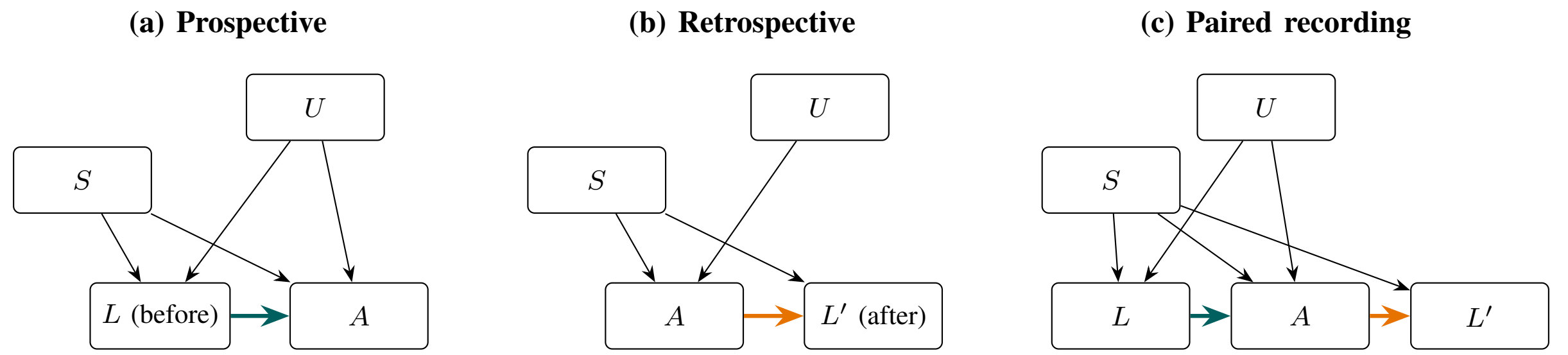
\begin{figure*}[t]
\centering
\begin{tikzpicture}[>=Stealth, x=1cm,y=1cm, font=\footnotesize,
  v/.style={draw,rounded corners=2pt,align=center,minimum width=1.42cm,minimum height=.68cm},
  title/.style={font=\small\bfseries}]
\node[title] at (2.7,3.35) {(a) Prospective};
\node[v] (Sa) at (1.2,1.75) {$S$};
\node[v] (Ua) at (3.6,2.5) {$U$};
\node[v] (La) at (2.0,.35) {$L$ (before)};
\node[v] (Aa) at (4.05,.35) {$A$};
\draw[->] (Sa)--(La); \draw[->] (Sa)--(Aa); \draw[->] (Ua)--(La); \draw[->] (Ua)--(Aa);
\draw[->,very thick,teal!75!black] (La)--(Aa);
\node[title] at (8.0,3.35) {(b) Retrospective};
\node[v] (Sb) at (6.5,1.75) {$S$};
\node[v] (Ub) at (8.8,2.5) {$U$};
\node[v] (Ab) at (7.3,.35) {$A$};
\node[v] (Lb) at (9.35,.35) {$L'$ (after)};
\draw[->] (Sb)--(Ab); \draw[->] (Sb)--(Lb); \draw[->] (Ub)--(Ab);
\draw[->,very thick,orange!90!black] (Ab)--(Lb);
\node[title] at (13.5,3.35) {(c) Paired recording};
\node[v] (Sc) at (11.8,1.75) {$S$};
\node[v] (Uc) at (13.4,2.5) {$U$};
\node[v] (Lc) at (11.9,.35) {$L$};
\node[v] (Ac) at (13.75,.35) {$A$};
\node[v] (Rc) at (15.6,.35) {$L'$};
\draw[->] (Sc)--(Lc); \draw[->] (Sc)--(Ac); \draw[->] (Sc)--(Rc);
\draw[->] (Uc)--(Lc); \draw[->] (Uc)--(Ac);
\draw[->,very thick,teal!75!black] (Lc)--(Ac);
\draw[->,very thick,orange!90!black] (Ac)--(Rc);
\end{tikzpicture}
\caption{Illustrative causal graphs, with observed pre-instruction context $S$, latent common causes $U$, prospective instruction $L$, recorded action $A$, and retrospective label $L'$. (a) A genuine request can influence the driver's subsequent action, but $U$ can confound the association. (b) A post-hoc label depends on the completed action and cannot affect that already-recorded action. (c) Our paired recordings provide both instruction types for the same action. Arrows depict possible dependencies, not an identified causal effect; some context and annotator-specific influences are omitted for clarity.}
\label{fig:casual_graphs}
\end{figure*}

\subsection{Why a Retrospective Label Can Predict an Action Without Causing It}
For an already-recorded human maneuver, intervening on a label written \emph{afterward} cannot alter that past maneuver. In the retrospective graph of Fig.~\ref{fig:casual_graphs}(b), the intervention thus satisfies
\begin{equation}
P(A\mid\doop(L'=\ell'),S)=P(A\mid S).
\label{eq:retro}
\end{equation}
This is a statement about the temporal data-generating process, not about the behavior of a VLA trained on these labels. Since the annotator may have observed $A$, the conditional distribution can still be predictive:
\begin{equation}
P(A\mid L'=\ell',S)\neq P(A\mid S)\quad\text{in general}.
\label{eq:retroassoc}
\end{equation}
Training on such correlations is not inherently invalid: hindsight instructions can supply meaningful goals for policy learning~\cite{andrychowicz2017hindsight,xiao2023robotic}. The unanswered empirical question is whether the learned policy follows new instructions robustly when they conflict with cues about the recorded action.

\subsection{Prospective Timing Makes an Effect Possible, Not Automatically Identified}
In a prospective process, the estimand for human instruction-following is the action distribution under an intervention on a request:
\begin{equation}
P(A\mid\doop(L=\ell),S).
\label{eq:prosestimand}
\end{equation}
Temporal precedence admits an $L\!\rightarrow\!A$ path, but does not establish that it is active for every instruction, nor that it can be identified from observational data. In particular, $U$ may affect the navigator's route instruction and the driver's maneuver; observed $S$ need not capture destination intent, traffic decisions, or compliance. Only under assumptions including sufficient confounder adjustment, consistency, and positivity may one equate the observational and interventional conditional distributions:
\begin{equation}
P(A\mid\doop(L=\ell),S=s)=P(A\mid L=\ell,S=s).
\label{eq:prosassumption}
\end{equation}
We \emph{do not assume} this equality holds automatically in our drives. Moreover, $S$ must be strictly pre-instruction for this interpretation; later frames can already reflect effects of the instruction. At later VLA prediction times, recent ego motion is a measured part of the evolving state, not necessarily a pre-treatment confounder. This distinction motivates our explicit intervention experiments on \emph{model inputs} rather than interpreting prospective timing alone as proof of causal human behavior.

\subsection{Instruction Interventions on a Frozen Driving Model}
For a fixed checkpoint with parameters $\theta$, define $M_\theta(S,\ell;\epsilon)$ as the predicted maneuver under instruction $\ell$ and model randomness $\epsilon$. Holding $S$ fixed and replacing $\ell$ makes a controlled \emph{input intervention} on the trained model. For requests $\ell\in\mathcal L_{\mathrm{legal}}(S)$, its instruction-conditioned counterfactual accuracy is
\begin{equation}
\mathrm{Acc}_{\mathrm{cf}}=\frac{1}{N}\sum_{i=1}^{N}\mathbf 1\big[M_\theta(S_i,\ell_i;\epsilon_i)=m(\ell_i)\big],
\label{eq:cfacc}
\end{equation}
where $m(\ell)$ is the maneuver requested by the language and the sum covers the alternative legal requests. The reported gain measures its difference from scoring the \emph{no-instruction} prediction against the same requested maneuvers:
\begin{equation}
\mathrm{Gain}_{\mathrm{cf}}=100\big(\mathrm{Acc}_{\mathrm{cf}}-\mathrm{Acc}_{\varnothing}\big)
\quad\text{percentage points}.
\label{eq:cfgain}
\end{equation}
Here the accuracies in Eq.~\ref{eq:cfgain} are fractions on $[0,1]$. A changed maneuver (Chg.) need not be correct, and stochastic sampling may cause changes even with identical instructions. These tests directly estimate a model's response to its instruction input \emph{at a frozen observed state}; they do not identify the causal effect of the historical instruction on the human driver or guarantee closed-loop compliance.

\subsection{What Paired Prospective and Retrospective Labels Let Us Test}
The same video can provide a prospective $L$ and multiple retrospective $L'_j$. Paired annotation comparisons can identify differences in wording, specificity, and temporal anchoring \emph{conditional on the same observed drive}, but cannot by themselves show that the retrospective mechanism caused a difference in driver behavior. Comparing separately fine-tuned policies requires holding training videos, sampling, model initialization, and compute budget fixed while varying the language and its assigned timing. Such a study estimates the effect of the \emph{training intervention on the model} under that experimental design. It does not imply the retrospective label caused the original maneuver. We lay out these experiments in Sec.~\ref{sec:planned}; their results remain pending.

\section{ADVLAT-Engine and Driving Data}
\label{sec:data}
\begin{figure}[t]
\centering
\includegraphics[width=\columnwidth]{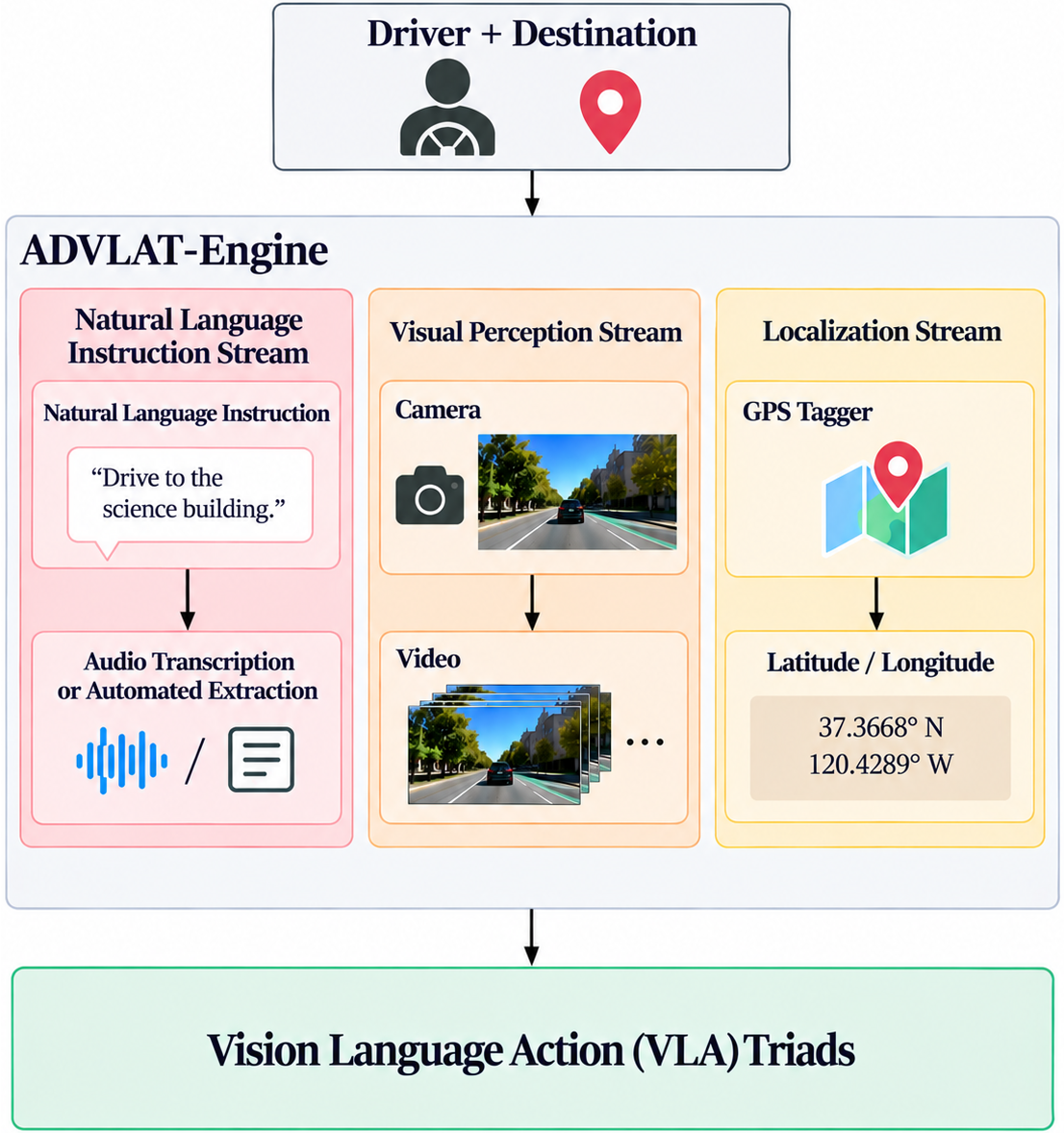}
\caption{ADVLAT-Engine synchronizes spoken navigation or passenger instructions, forward camera observations, and vehicle localization to form instruction--vision--trajectory training examples.}
\label{fig:sample_advlat}
\end{figure}

ADVLAT-Engine aligns visual observations, spoken instructions, and vehicle localization on a common timeline. Figure~\ref{fig:sample_advlat} shows the three synchronized streams. Our study uses a forward-facing Intel RealSense D455F camera (1280$\times$720 at 30~fps, with IMU), a 10~Hz u-blox GNSS receiver, a passenger microphone, and Google Maps audio captured digitally from an Android phone. Sensor events use a shared monotonic clock; video frames are associated with GNSS-disciplined UTC. Whisper~\cite{radford2023robust} provides transcriptions, reviewed after each drive. An operator marks maneuver completion.

An instruction becomes active when it has been \emph{fully spoken} and stays active until maneuver completion or a newer instruction from that same source. No instruction is active between completion and the next prompt. Thus a sample contains the observations and instruction actually available to the driver at the prediction time. Future trajectories are generated from subsequent GNSS positions with heading estimated using a gyroscope/GNSS complementary filter. This differs from creating an instruction after inspecting the future video.

\subsection{Navigation-App Comparison}
\label{sec:apps}
We replayed 48 routes across four road types and three metropolitan areas through Apple Maps, Google Maps, and Waze. Each app received the same simulated GNSS track. Prompts were transcribed and categorized into eight potentially overlapping referentiality classes: turn, road name, distance, static object, lane, traffic light, location name, and cardinal direction. Paired analysis uses the 170 maneuvers announced by all three apps (Table~\ref{tab:apps}). Cochran's Q was used for category comparisons with Bonferroni correction across eight classes.

\begin{table}[t]
\centering\footnotesize
\caption{Navigation-app prompts on 170 shared maneuvers (\%); median first-prompt leads are measured at replayed speed.}
\label{tab:apps}
\setlength{\tabcolsep}{3.5pt}
\begin{tabular}{lrrr}\toprule
 & Apple & Google & Waze\\\midrule
Turn &95.9&92.9&97.6\\
Road name &91.8&94.7&96.5\\
Distance &48.2&48.2&78.8\\
Static object &64.7&67.1&38.2\\
Lane &18.8&20.6&25.3\\
Light &17.1&4.7&0.6\\
Location name &7.6&7.1&0.0\\
Cardinal &0.0&0.6&0.0\\\midrule
First lead (m) &468&259&369\\
First lead (s) &41.6&25.3&33.5\\\bottomrule
\end{tabular}
\end{table}

Waze provides distances for about 79\% of shared maneuvers, versus about 48\% for Apple and Google; Apple and Google reference static objects more often. These differences change the language available for training. For example, acting on a distance requires estimating how far away the turn lies, while acting on a stop-sign reference requires identifying the sign and relating it to the appropriate maneuver. We do not infer that training on one app necessarily teaches these skills better; cross-app model training remains untested.

\subsection{Prospective Drive Collections}
We recorded 60 Google Maps-guided drives (84 minutes in daylight and 88 at night), comprising 406 app instructions; 54 drives provided 17,687 training samples and six were held out for validation. We separately recorded 28 passenger drives (62 minutes, 135 passenger instructions), with Google Maps muted but its 161 prompts logged simultaneously. The passenger drives provide a test set for generalization from app prompts to human language. Data come from one city and one passenger.

In randomized-route drives, legal maneuvers were selected randomly at intersections, with text-to-speech instructions issued after clearing the previous intersection. The resulting 321 encounters span 117 distinct intersections: 67 allow left/right, 58 right/straight, 51 left/straight, and 145 all three. Logged legal options permit counterfactual prompts that do not match the recorded turn.

\begin{figure*}[t]
\centering
\includegraphics[width=\textwidth]{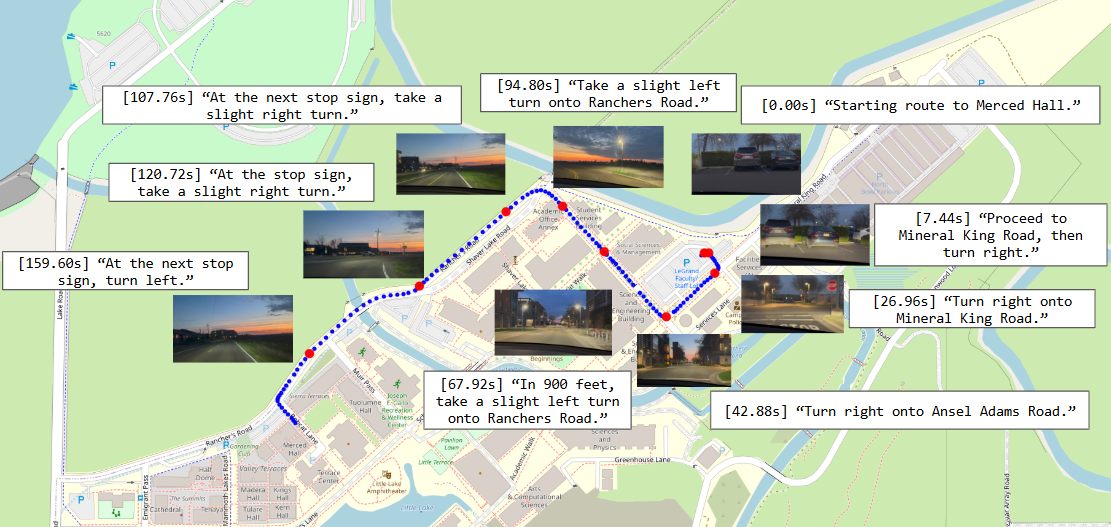}
\caption{Example navigation-app recording with a GNSS route, speech-completion positions, transcribed prompts, and associated camera frames. The original prompts were spoken during the drive, rather than written after viewing its outcome.}
\label{fig:ucmerced_example_run}
\end{figure*}

\subsection{Four-Person Retrospective Annotation Collection}
\label{sec:retrodata}
Four people also annotated the recorded driving videos retrospectively. Their annotations constitute a separate human-authored source for comparison with instructions actually spoken during the drives. Because these labels were produced after recording, their content may reflect the completed maneuver and the available video context. The current manuscript does \emph{not} report annotation agreement, temporal alignment, instruction similarity, or model-training results from this four-person set: these analyses have not been supplied and are specified in Sec.~\ref{sec:planned}. 

\section{How Passenger Instructions Differ from App Prompts}
\label{sec:human}
To understand the domain shift from apps to people, we matched the 87 maneuvers announced by both the passenger and Google Maps using their common completion marker. Table~\ref{tab:human} summarizes language and timing differences. Content comparisons use exact McNemar tests with Bonferroni correction over 12 classes; the displayed descriptive percentages do not by themselves establish a population-wide difference, since the study involves one passenger.

\begin{table}[t]
\centering\footnotesize
\caption{Passenger and Google Maps prompts on the same 87 maneuvers (\% unless indicated).}
\label{tab:human}
\setlength{\tabcolsep}{4pt}
\begin{tabular}{lrr}\toprule
 &Passenger&Google Maps\\\midrule
Road name &1.1&93.1\\
Distance &0.0&34.5\\
Static object &52.9&67.8\\
Deictic (``here'', ``now'') &43.7&1.1\\
Continue/follow road &36.8&1.1\\
None of eight app classes &20.7&0.0\\\midrule
First lead (s), median &22.8&24.7\\
First lead (m), median &176&196\\
\bottomrule\end{tabular}
\end{table}

Both sources usually specify the turn, yet the passenger almost never names the road and never states a distance. Instead, deictic language such as ``right here'' appears in nearly 44\% of matched maneuvers. The passenger's first instruction arrives at a median 22.8 seconds before maneuver completion versus 24.7 seconds for Google Maps; the paired median difference is 0.3 seconds (95\% CI $[-1.1,2.0]$). Thus both supply prospectively timed supervision, but with substantially different linguistic forms. These results motivate testing whether app-trained VLAs can generalize to passenger directions.

\section{Benchmarking Driving VLAs}
\label{sec:benchmark}
\subsection{Models and Training}
We evaluate Alpamayo~1.5 (10B), Alpamayo~2 Super (34B), Qwen-Drive (4B), and Impromptu~VLA (7B). Each is adapted to a single front-facing camera; Alpamayo~2 Super receives six black images for unavailable cameras. Qwen-Drive receives the full text instruction, but its discrete navigation command is provided only for bare directives on the passenger test (otherwise ``unknown''). Impromptu~VLA was released without instruction conditioning and provides an instructive negative control. The models are fine-tuned on app drives for two epochs using their respective training objectives; Alpamayo uses its two-stage recipe, while Qwen-Drive uses a reconstructed flow-matching training loss and direct-planning mode because training code was not released. Changes in Qwen-Drive's inference mode complicate before/after comparison.

Passenger samples are extracted at 0.5-second intervals with at least three seconds of future trajectory. We evaluate 5,012 instructed timestamps both with and without the passenger's instruction; 1,897 additional timestamps have no instruction. We measure average and final displacement errors (ADE, FDE) at a common 3-second horizon, and left/right turn accuracy on samples where the recorded path changes heading by more than 30 degrees between 2.5 and 3 seconds. The paired effect is ADE(with instruction) minus ADE(without), with confidence intervals bootstrapped across the 135 passenger instructions. A speed/yaw-rate extrapolation serves as a language-free baseline.

\begin{table*}[t]
\caption{Passenger-drive test, 3-s horizon. ADE and FDE in meters; Effect = instructed minus uninstructed ADE (negative indicates improvement), with 95\% CI. Turn is left/right accuracy (\%).}
\label{tab:main}
\centering\footnotesize\setlength{\tabcolsep}{6pt}
\begin{tabular}{lrrlrr}\toprule
& \multicolumn{2}{c}{ADE $\downarrow$} & Effect $\downarrow$ & FDE $\downarrow$ & Turn $\uparrow$ \\
\cmidrule(lr){2-3}
Model & Instr. & No instr. & Difference [95\% CI] & Instr. & L / R\\\midrule
Ego extrap. & \multicolumn{2}{c}{1.331} & -- & 3.16 & 60 / 54\\\midrule
Alpamayo 1.5 &1.887&1.863&$+0.024$ [$+0.001,+0.052$]&4.38&30 / 6\\
\quad fine-tuned&0.775&0.772&$+0.003$ [$-0.010,+0.018$]&1.78&83 / 83\\
Alpamayo 2 Super&1.381&1.391&$-0.010$ [$-0.026,+0.008$]&3.19&90 / 88\\
\quad fine-tuned&0.693&0.686&$+0.007$ [$-0.010,+0.029$]&1.56&\textbf{92} / 84\\
Qwen-Drive&0.787&0.833&$-0.046$ [$-0.062,-0.031$]&1.76&75 / 69\\
\quad fine-tuned&0.609&0.661&$\mathbf{-0.052}$ [$-0.082,-0.026$]&\textbf{1.36}&79 / \textbf{90}\\
Impromptu VLA&1.011&1.003&$+0.008$ [$+0.001,+0.018$]&2.37&0 / 1\\
\quad fine-tuned&\textbf{0.554}&0.579&$-0.025$ [$-0.032,-0.017$]&1.37&81 / 80\\\bottomrule
\end{tabular}
\end{table*}

\subsection{Counterfactual Intersection Evaluation}
\label{sec:intersection}
At each randomized-route encounter, we freeze observations 15, 5, and 0 meters from a hand-marked stop line. We query every legal maneuver plus a no-instruction condition while holding camera frames and ego history fixed. \emph{Counterfactual} prompts ask for legal maneuvers other than the one executed by the driver. We use each model's native prediction horizon (3, 5, or 6.4 seconds) to make turns visible when possible. At 0 m we also construct a \emph{parked} condition by repeating the current frame across the visual history and setting ego-motion history to stationary. This changes both histories, so it cannot isolate ego motion alone.

Maneuvers are classified from the last 0.5 seconds of predicted heading, with left/right beyond $\pm30^\circ$ and straight otherwise. \emph{Acc.} is counterfactual accuracy, \emph{Gain} is improvement over the no-instruction prediction scored against the same requested maneuver (percentage points), and \emph{Chg.} is the percentage whose predicted maneuver class changes regardless of correctness. \emph{Steer} is the mean \emph{signed} final-heading difference between left and right prompts, with no absolute value before or after averaging. \emph{Rest} is the fraction of parked predictions remaining stationary. \emph{Ceil.} is a horizon-visibility diagnostic: the fraction of recorded trajectories in which the \emph{executed} maneuver appears within that model's horizon; it is not a strict ceiling on counterfactual accuracy.

\begin{table*}[t]
\caption{Counterfactual results at the hand-marked stop line. Accuracy in \%, Gain in percentage points, Steer in degrees (signed left-minus-right). At 0 m, \emph{driven} uses the recorded approach and \emph{parked} uses repeated frames and stationary ego history. Bold highlights maximum accuracy/gain within a condition, including ties.}
\label{tab:counter}
\centering\scriptsize\setlength{\tabcolsep}{3pt}
\begin{tabular}{lrrrrrrrrrr}\toprule
&\multicolumn{2}{c}{15 m}&\multicolumn{2}{c}{5 m}&\multicolumn{3}{c}{0 m driven}&\multicolumn{3}{c}{0 m parked}\\
\cmidrule(lr){2-3}\cmidrule(lr){4-5}\cmidrule(lr){6-8}\cmidrule(lr){9-11}
Model &Acc.&Gain&Acc.&Gain&Acc.&Gain&Steer&Acc.&Gain&Steer\\\midrule
Alpamayo 1.5&31&+2&27&+2&23&+1&1&27&0&0\\
\quad fine-tuned&24&+4&13&+3&1&0&1&19&0&5\\
Alpamayo 2 Super&32&+8&10&+2&2&0&0&15&$-1$&0\\
\quad fine-tuned&55&+31&25&+18&4&+3&7&34&+15&40\\
Qwen-Drive, sentence&43&+19&34&+21&\textbf{93}&\textbf{+86}&151&21&+15&42\\
\quad fine-tuned&42&+17&35&+22&45&+42&104&\textbf{59}&\textbf{+33}&97\\
Qwen-Drive, + command&\textbf{78}&\textbf{+54}&\textbf{89}&\textbf{+77}&\textbf{93}&\textbf{+86}&151&21&+15&42\\
Impromptu VLA&29&0&27&0&29&0&$-1$&0&0&--\\
\quad fine-tuned&29&0&14&$-1$&7&+4&12&0&0&--\\\bottomrule
\end{tabular}
\end{table*}

\begin{figure*}[t]
\centering
\includegraphics[width=\textwidth]{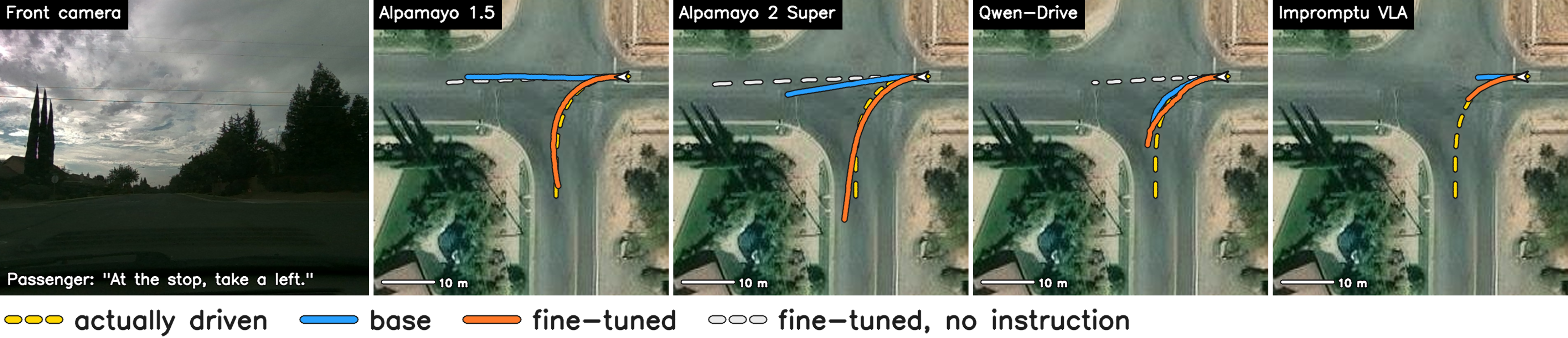}
\caption{Passenger-drive example: observed trajectory and pretrained/fine-tuned model predictions compared with fine-tuned predictions after removing the passenger instruction. A single trajectory illustration complements, but does not replace, paired error and counterfactual instruction tests.}
\label{fig:passenger_bev}
\end{figure*}

\section{Results: What App-Drive Training Establishes}
\label{sec:results}
\textbf{Off-the-shelf instruction use.} On passenger drives (illustrated in Fig.~\ref{fig:passenger_bev}), Qwen-Drive is the only pretrained model with a clear reduction in ADE when given the instruction: $-0.046$ m (95\% CI $[-0.062,-0.031]$), about 6\%. The other models' paired ADE changes have magnitude at most 0.024 m. Alpamayo~2 Super predicts observed turns accurately (90/88\%) while showing little paired ADE response; Impromptu~VLA achieves relatively low overall ADE while almost never predicting turns. Overall displacement error includes driving before a turn enters the 3-second horizon, so it can conceal poor maneuver prediction or weak instruction reliance.

\textbf{Transfer after fine-tuning.} Fine-tuning on app drives reduces ADE by 23--59\% across models, with much of the improvement retained when the passenger's instruction is removed. Left/right turn accuracy increases from 30/6\% to 83/83\% for Alpamayo~1.5, and from 0/1\% to 81/80\% for Impromptu~VLA. These changes are consistent with improved maneuver prediction but do not establish that the navigation \emph{text}, rather than associated imagery, ego history, or environment adaptation, caused the gain. Qwen-Drive and Impromptu~VLA show negative paired ADE effects after fine-tuning ($-0.052$ and $-0.025$ m); both fine-tuned Alpamayo models have near-zero mean paired effects.

\textbf{Counterfactual responses.} At 15 and 5 meters, pretrained Qwen-Drive follows 78/89\% of alternative legal maneuvers with an explicit discrete command, but 43/34\% when given the sentence alone. Fine-tuned Alpamayo~2 Super shows a rise in counterfactual gain from +8/+2 to +31/+18 points at those distances even though its paired passenger ADE effect is small. At 0 m as driven, its counterfactual accuracy is only 4\%, versus 45\% for fine-tuned Qwen-Drive and 93\% for command-conditioned pretrained Qwen-Drive. Hence a model may learn some instruction sensitivity without following conflicting instructions reliably throughout an approach.

\textbf{History dependence.} Fine-tuned Alpamayo~2 Super's 0-m gain rises from +3 points as driven to +15 parked, and signed steering shift rises from $7^\circ$ to $40^\circ$. Fine-tuned Qwen-Drive retains a substantial gain (+42 driven, +33 parked), whereas pretrained command-conditioned Qwen-Drive drops from +86 to +15. These interventions show that instruction response depends on the recorded context. Because the parked input changes both visual and ego-motion histories and is out of distribution, the result is not an ego-only causal ablation.

These completed experiments establish an important distinction for the next stage of this study: app recordings can provide effective supervision for predicting driving trajectories, but trajectory gains alone do not show that models are using the language. Whether prospective training data leads to more robust instruction reliance than retrospective labels \emph{for the same recorded videos} remains an open empirical question.

\section{Planned Prospective--Retrospective Comparisons}
\label{sec:planned}
The four-annotator retrospective collection enables controlled experiments that have \emph{not yet been evaluated}. In this section, we specify methods in which can be use to evaluated the difference in prospective and retrospective annotations. 

\subsection{RQ1: How Do Annotations Differ on the Same Maneuver?}
Match each retrospective label to its corresponding recorded maneuver, and compare it with the prospective app instruction and, where available, the passenger's instruction. Use the same eight reference classes as the app analysis, augmented by deictic expressions, references to transient objects, and whether the instruction specifies a maneuver. Report class prevalence, median length, and within-maneuver agreement. For the four retrospective annotators, report inter-annotator agreement separately for maneuver labels and reference classes, using a statistic suited to multiple annotators and multi-label categories (e.g., per-class Fleiss' kappa or Krippendorff's alpha~\cite{krippendorff2011agreement}, alongside raw agreement). This tests whether retrospective language is consistent and whether it differs systematically from what was said prospectively. No direction or magnitude of the difference should be assumed in advance.

\subsection{RQ2: Does Retrospective Language Reveal the Outcome?}
Because annotators may observe a completed turn, test whether their labels specify the executed maneuver more often than the prospective language and whether they mention cues visible only later. Compare the fraction of labels consistent with the recorded trajectory and classify outcome-derived wording separately from scene references available at the instruction time. A useful audit presents annotators with truncated video ending at the original instruction time and asks whether each label could have been justified without seeing the future. If such an audit is conducted, report its protocol and agreement rather than infer hindsight from wording alone.

\subsection{RQ3: What Does It Mean to Align a Retrospective Label in Time?}
Prospective prompts have genuine speech-completion timestamps, so their lead-time distribution is measurable. Retrospective labels have no naturally observed issue time. Compare plausible, explicitly documented anchor policies: (i) the original prospective instruction time for the paired maneuver, (ii) a fixed lead time before completion, and (iii) a maneuver-proximal time. Hold language and video fixed when changing only the anchor. Report how these choices affect data availability and the number of windows in which the turn is visible within 3, 5, and 6.4 seconds. The artificial anchors must not be described as when the retrospective labels were actually spoken.

\subsection{RQ4: Which Supervision Better Trains Instruction Use?}
Fine-tune identical checkpoints using the \emph{same driving videos, sample counts, training budgets, and splits}, varying only the source and timing of language. Compare prospective app text, retrospective text from each annotator (and an agreed or sampled multi-annotator variant), and a no-instruction training condition. Retain the same language-free intervals across arms. Evaluate on held-out passenger instructions, reporting ADE/FDE, paired instruction effects, and the already defined counterfactual accuracy and gain. Because annotator wording may differ from passenger wording, also measure basic instruction-style overlap to distinguish source mismatch from prospective timing. Repeated seeds and confidence intervals clustered by drive or intersection are important; sampling frames as independent observations would overstate precision.

\subsection{RQ5: Can Model Changes Be Attributed to the Instruction?}
Use paired no-instruction and alternative-legal-instruction queries from identical observations. In addition to the existing parked experiment, separately change ego-motion history while holding video history fixed, and vice versa, to estimate their individual influence. For stochastic models, run repeated controlled seeds to estimate changes caused by sampling rather than language. A closed-loop experiment can then test whether a changed instruction leads to a sustained alternative maneuver once the resulting observations differ from the original recording. These experiments extend, rather than invalidate, the current real-video open-loop results.

\section{Discussion and Limitations}
\label{sec:discussion}
Our work separates three questions that are often conflated: whether instruction language can be collected at scale, whether training on instruction-associated drives improves trajectory prediction, and whether a model's output changes \emph{correctly} when its instruction changes. The navigation-app and passenger recordings address the first question; the completed fine-tuning and counterfactual experiments address the latter two. The four-person retrospective annotations provide a matched dataset for asking whether the \emph{construction} of training language influences those outcomes, without yet establishing an answer.

The open-loop counterfactual protocol offers a controlled test of alternative instructions at identical recorded moments. Its limitation is that a model's subsequent predictions remain conditioned on the original approach, even when asked to perform a different maneuver. Closed-loop evaluation could extend the findings by allowing the vehicle to follow an alternative and observe the consequences. A further limitation is that our on-road passenger observations involve one city and one passenger. Model adaptations also matter: Alpamayo~2 Super receives six black cameras instead of its native multi-camera input, Qwen-Drive switches from reasoning to direct planning during fine-tuning, and Impromptu~VLA's 3-second horizon may end before a recorded turn becomes visible. The horizon-visibility diagnostic (Ceil.) helps interpret that last limitation but does not measure the feasibility of every alternative turn.

Neither source timing nor the causal-diagram distinction proves that a particular recorded instruction caused a particular human maneuver. Route selection, traffic, driver intent, and instruction compliance must be considered. Further, retrospective labels can still provide effective instruction-conditioned training even though they were produced after the action. The pending matched training experiments, especially the no-instruction control, are needed before attributing performance differences to prospective supervision itself. Collection and release of speech, faces, and trajectories also require consent and appropriate privacy safeguards; no release is asserted here until those steps and accessible artifacts are in place.

\section{Conclusion}
Navigation apps provide a scalable source of naturally timed language paired with road scenes and subsequent trajectories, while passengers supply more ambiguous, scene-dependent directions. Across four driving VLAs, app-drive fine-tuning improves trajectory prediction, but models vary in whether they follow the instruction rather than the maneuver implied by the recorded scene. Counterfactual evaluation reveals differences that ordinary displacement error can miss. The prospective-versus-retrospective distinction clarifies a further question: how much does a model's learned behavior depend on whether training language was spoken before the action or written after observing it? Our four-annotator collection and planned matched comparisons make that question testable. Answering it can inform data collection for passenger-directed autonomous vehicles, including requests to move toward a particular curb or entrance rather than merely reach a map destination.

\bibliographystyle{IEEEtran}
\bibliography{refs}
\end{document}